\documentclass[11pt,a4paper]{article}
\usepackage[hyperref]{emnlp2020}
\usepackage{times}
\usepackage{latexsym}

\usepackage{subcaption}
\usepackage{pgfplots}
\usepackage{tikz}
\usepackage{latexsym}

\usepackage{amsmath}
\usepackage{float}
\usepackage{adjustbox}
\usepackage{multirow}

\DeclareUnicodeCharacter{01B0}{u}
\DeclareUnicodeCharacter{01A1}{o}
\DeclareUnicodeCharacter{0300}{}

\makeatletter
\def\new@fontshape{}
\makeatother
\let\expold\exp
\usepackage{gb4e}
\noautomath 
\let\exp\expold
\makeatletter
\newcommand*\iftodonotes{\if@todonotes@disabled\expandafter\@secondoftwo\else\expandafter\@firstoftwo\fi}  
\makeatother

\usepackage{microtype}
\usepackage{amsmath}
\usepackage{amsfonts}
\usepackage{amsthm}
\usepackage{amssymb}
\usepackage{enumerate}
\usepackage{xspace}

\everypar{\looseness=-1}

\aclfinalcopy 

\usepackage{cleveref}
\crefname{section}{\S}{\S\S}
\Crefname{section}{\S}{\S\S}
\crefname{table}{Tab.}{Tabs.}
\crefname{figure}{Fig.}{Figs.}
\crefname{algorithm}{Algorithm}{Algorithms}
\crefname{equation}{eq.}{eqs.}
\crefname{appendix}{App.}{Apps.}
\crefname{thm}{Theorem}{Theorems}
\crefname{prop}{Proposition}{Propositions}
\crefname{cor}{Corollary}{Corollaries}
\crefname{observation}{Observation}{Observations}
\crefname{assumption}{Assumption}{Assumptions}
\crefformat{section}{\S#2#1#3}
\crefformat{xnumi}{#2(#1)#3}
\crefmultiformat{xnumi}{#2(#1)#3}{ and~#2(#1)#3}{, #2(#1)#3}{, and~#2(#1)#3}
\crefrangeformat{xnumi}{#3(#1)#4--#5(#2)#6}
\crefformat{xnumii}{#2(#1)#3}
\crefformat{xnumiii}{#2(#1)#3}
\crefformat{xnumiv}{#2(#1)#3}

\usepackage{booktabs}
\usepackage{todonotes}

\newcommand{\vpi}{{\boldsymbol \pi}} 
\newcommand{\va}{{\mathbf{a}}}
\newcommand{\ve}{{\mathbf{e}}}

\newcommand{\score}{\mathrm{score}}

\newcommand{\calC}{\mathcal{C}}
\newcommand{\Sk}{S_k}
\newcommand{\vc}{\mathbf{c}}

\newcommand{\R}{\mathbb{R}}

\newcommand{\softmax}{\mathrm{softmax}}
\newcommand{\bV}{\mathbf{V}}
\newcommand{\bW}{\mathbf{W}}

\title{Investigating Cross-Linguistic Adjective Ordering Tendencies\\with a Latent-Variable Model}

\usepackage{tipa}
\usepackage{emoji}
\usepackage[utf8]{inputenc}
\newcommand{\ucambridge}{\emoji[twitter]{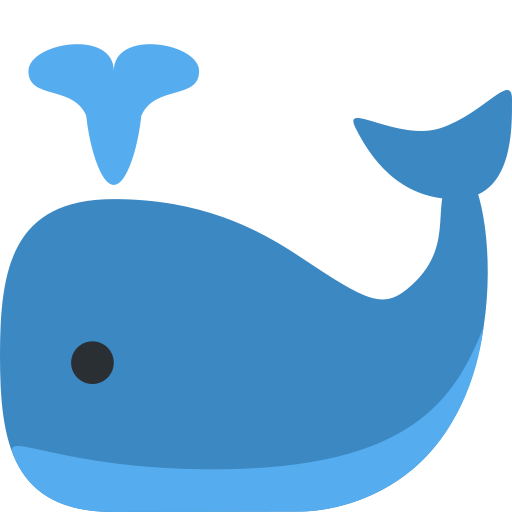}}
\newcommand{\ethz}{\emoji[twitter]{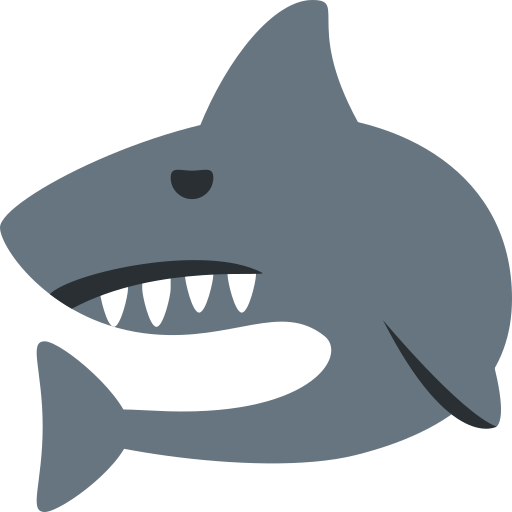}}

\author{Jun Yen Leung$^{\ucambridge}$ Guy Emerson$^{\ucambridge}$ Ryan Cotterell$^{\ucambridge, \ethz}$\\
  $^{\ucambridge}$University of Cambridge~\;~
  $^{\ethz}$ETH Z\"{u}rich \\
  \texttt{junyenle@gmail.com},~\;~ \texttt{gete2@cam.ac.uk},~\;~ \texttt{ryan.cotterell@inf.ethz.ch}
}

\date{}

\begin{document}
\pgfplotsset{compat=1.16}
\maketitle
\begin{abstract}

Across languages, multiple consecutive adjectives modifying a noun (e.g.~``the big red dog'') follow certain unmarked ordering rules. While explanatory accounts have been put forward, much of the work done in this area has relied primarily on the intuitive judgment of native speakers, rather than on corpus data. We present the first purely corpus-driven model of multi-lingual adjective ordering in the form of a latent-variable model that can accurately order adjectives across 24 different languages, even when the training and testing languages are different. We utilize this novel statistical model to provide strong converging evidence for the existence of universal, cross-linguistic, hierarchical adjective ordering tendencies.
\end{abstract}

\section{Introduction}
Most native speakers of a language would agree that certain adjective orderings are preferable to others. 
For instance, in English, ``the big red dog'' sounds natural while ``the red big dog'' sounds very awkward. 
Similar ordering preferences have been found to apply universally across the languages in the world: for example, the adjective for  ``big'' in most languages tends to be farther away from the noun, syntactically, than ``red.'' 
For an overview of these phenomena, see \newcite{cinquebook}.

There are many explanatory accounts of cross-linguistic adjective ordering in the linguistics literature, the most popular being hierarchical tendencies based on semantic categories of adjectives \cite{dixon, sproat, cinque1994, cinquebook}. For instance, \newcite{sproat} and \newcite{cinquebook} note that adjectives describing $\textsc{size}$ tend to be placed further from the noun than those describing $\textsc{color}$ in most languages. However, most of these studies have relied primarily on the judgment of native speakers rather than on corpus data, and those corpus-based models that do exist have focused exclusively on English \cite{shaw-hatzivassiloglou-1999-ordering, malouf, wuff, mitchell, dunlop, mitchelll, hill, scontras, hahn, futrell-etal-2020-determines}. In this paper, we make use of tools and techniques from statistical modeling to provide strong converging evidence supporting a hierarchical theory of cross-linguistic adjective ordering. 

Specifically, we present a novel interpretable, multi-lingual, latent-variable model of adjective ordering that directly enforces a hierarchy of semantic classes and is trained entirely using corpus data.
We empirically show that our model accurately orders adjectives across 24 different languages, even when tested on languages that it has not been trained on.
In doing so, we demonstrate the existence of universal, cross-linguistic, hierarchical tendencies in adjective ordering.

\section{Adjective Ordering}
Consider the following English phrases, taken from \newcite{teodorescu2006adjective}:
\begin{exe}
    \ex\label{ex:purse} A beautiful small black purse
    \ex \begin{xlist}
        \ex \# A beautiful black small purse\footnote{\# denotes an infelicitous phrase}
        \ex \# A small beautiful black purse
        \ex \# A small black beautiful purse
    \end{xlist}
\end{exe}
\noindent None of these phrases are ungrammatical, yet most native English speakers would contend that only \cref{ex:purse} is correct in most contexts.
Further complicating the phenomenon, there are many unmarked cases where ordering rules can be broken without hurting correctness. For example, now consider:
\begin{exe}
    \ex\label{ex:bear} A brown Chinese bear
    \ex\label{ex:brownbear} A Chinese [brown bear]\footnote{[ ] denotes an adjective-noun collocate}
\end{exe}
\noindent Here, \cref{ex:bear} presents the most natural ordering of ``brown'' and ``Chinese'' (to illustrate this, substitute ``bear'' with ``house''), but \cref{ex:brownbear} is also correct because a ``brown bear'' is an adjective--noun collocate.
For a more detailed discussion on adjective ordering exceptions, see \newcite{teodorescu2006adjective}.

\subsection{Common Theories}
All adjective ordering theories put adjectives on a scale. What differentiates them is the granularity of that scale and the metric used to rank adjectives. This section describes the most notable theories, which appeal to a hierarchy of semantic classes, inherentness,
modification strength,
and subjectivity. 
We adopt the hierarchical approach in this paper because it is more general and so allows a closer fit to the data.  While the more functional explanations (i.e.~inherentness, modification strength, and subjectivity) might allow us to derive a hierarchy from something more fundamental, current theories only appear to account for a portion of adjective ordering preferences.

\paragraph{Hierarchical theories.} Hierarchical theories of adjective ordering posit that each adjective belongs to a class of semantically similar adjectives, and that these classes follow a rigid order. 
Several theories describing how prenominal adjective classes are ordered have been suggested, most famously
\newcite{cinquebook}'s: $\textsc{value}$ $\rightarrow$ $\textsc{size}$ $\rightarrow$ $\textsc{shape}$ $\rightarrow$ $\textsc{color}$ $\rightarrow$ $\textsc{provenance}$.
\newcite{dixon} observes that postnominal adjectives follow the opposite order as do prenominal ones.
To illustrate, consider the following phrase in both English and Spanish:
\begin{exe}
    \ex\label{ex:eng-shirt} An ugly black shirt
    \ex\label{ex:spa-shirt}\gll
        Una camisa negra fea\\
        \textit{a} \textit{shirt} \textit{black} \textit{ugly}\\
\end{exe}

\paragraph{Inherentness.}
The inherentness theory \cite{whorf} posits that adjectives fall into two broad categories: adjectives that describe inherent properties of nouns---such as color, material, physical state, provenance, breed, nationality, function, use, etc.---and adjectives that describe non-inherent properties, and that inherent adjectives are usually placed closer to the noun than non-inherent ones.

\paragraph{Modification strength.} \newcite{vecchi} apply a compositional distributional semantics approach to studying English adjective--adjective--noun phrases, and note that in correctly ordered phrases, the adjective closer to the noun contributes more to the meaning of the phrase than does the adjective further from the noun. For instance, ``different architectural style'' is more similar to ``architectural style'' than it is to ``different style''.

\paragraph{Subjectivity.} The subjectivity theory \cite{hill, scontras, hahn} ranks adjectives by subjectivity on a continuous  scale and posits that the less subjective an adjective is, the closer it should be placed to the noun.

\subsection{Binomial Ordering}
A closely related phenomenon to adjective ordering is binomial ordering. Binomials are pairs of words joined by a conjunction, such as ``salt and pepper'' or ``ball and chain''. Adjective ordering and binomial ordering have been studied in similar ways, and have in many cases been found to behave similarly \cite{benor, copestake, ivanova}.

\section{A Latent-Variable Model}
A natural mathematical formalization of adjective ordering is as a latent-variable model. A latent-variable model relates a set of observable variables
to a set of unobservable (latent) ones. Here, we observe how adjectives are ordered in corpus data and from this infer an ordered set of latent adjective classes. This allows us to determine the ordering of an arbitrary set of adjectives by referencing their class memberships and the class order.

Like other latent-variable models,
such as latent semantic analysis \cite{dumais}
and latent Dirichlet allocation \cite{blei},
our model aims to fit the data
using a lower-dimensional space.
In particular, the number of adjective classes
is much smaller than the size of the vocabulary
or the size of the pre-trained adjective embeddings.

\subsection{Ordering English Adjectives}\label{englishmodel}
Consider an English noun phrase where $k$ unique adjectives $\va = \{a_1, \ldots, a_k\}$ modify a noun $n$, and ${k \ge 2}$. Let $\calC$ be an ordered set of latent adjective classes labeled $[1, 2, \ldots,|\calC|]$ and let $d$ be the dimensionality of our pre-trained word embedding vectors $\ve(\cdot)$. Our goal is to simultaneously learn a mapping $\bV \in \R^{d \times |\calC|}$ from adjective embeddings to latent classes and learn an interaction matrix $\mathbf{W} \in \R^{|\calC| \times |\calC|}$ which reflects the preferred ordering of those classes.

We develop a probabilistic model of each of
the $k!$~possible permutations $\vpi$ of~$\va$ as in \cref{eqn:a},
which factorizes the distribution in terms of
the latent classes~$\vc$
(a $k$-length tuple of class labels,
one per adjective in the permutation).
The $i^{\text{th}}$ class,~$c_i$,
denotes the class assigned to
the $i^\text{th}$ adjective,~$a_i$.
\begin{equation}\label{eqn:a}
    p(\vpi \mid \va) = \sum_{\vc \in \calC^k} p(\vpi \mid \vc)\,\prod_{i=1}^k p(c_i \mid a_i) 
\end{equation}

Given latent classes, the distribution over permutations
is given in \cref{eqn:b},
using the scoring function in \cref{eqn:c},
where $\pi_i$ indexes the adjective in the $i^{\text{th}}$ position of the permutation,
and so $c_{\pi_i}$ is the latent class in the $i^{\text{th}}$ position.
Thus, \cref{eqn:c} sums the ordering preference scores between each consecutive pair of adjective classes in the permutation, using the pairwise preferences in~$\mathbf{W}$.
Using these scores, \cref{eqn:b} produces a distribution,
normalizing over the set of all permutations~$\Sk$:
\begin{align}
    p(\vpi \mid \vc) &= \frac{\exp\score(\vpi, \vc)}{\sum_{\vpi' \in \Sk} \exp  \score(\vpi', \vc)} \label{eqn:b} \\
    \score(\vpi, \vc) &= \sum_{i=1}^{k-1} W_{c_{\pi_i},c_{\pi_{i+1}}}
    \label{eqn:c}
\end{align}

Finally, the distribution over latent classes is obtained with~$\bV$,
making use of a pre-trained embedding~$\ve(a_i)$ for each adjective:
\begin{equation}\label{eqn:d}
    p(c_i \mid a_i) = \softmax\left(\bV\,\ve(a_i)\right)_{c_i}
\end{equation}

To summarize, we compute the probability of each permutation
by considering all possible assignments of latent classes.
The probability of a permutation is a weighted sum (\cref{eqn:a})
of normalized scores (\cref{eqn:b,eqn:c}, using $\bW$),
weighted according to the likelihood of the latent classes
(\cref{eqn:d}, using $\bV$).
Both $\mathbf{W}$ and $\bV$ are learned through batch gradient descent.

To predict an ordering, we enumerate all permutations of $\va$, compute their probabilities as described, and pick the highest scoring one.

\subsection{Enforcing a Total Ordering}
Hierarchical theories imply a total ordering of adjective classes. This means that the class order is antisymmetric, transitive, and a connex relation. While it is likely that our model learns a (predominantly) total ordering, we cannot be absolutely sure that it does. To remedy this, we enforce a total ordering of categories by modifying our model such that $\mathbf{W}$ is no longer learned, but is instead fixed as a matrix with ones above the diagonal and zeroes elsewhere. We will refer to this as an off-upper-triangular matrix. To illustrate how this enforces a total ordering, recall that each element $W_{ij}$ of $\mathbf{W}$ represents a preference for ordering class $i$ before class $j$. Then, given a $|\calC|\times|\calC|$ off-upper-triangular matrix of ones and zeroes:
$$
\begin{bmatrix} 
0 & 1 & 1 & \ldots & 1 \\
0 & 0 & 1 & \ldots & 1\\
0 & 0 & 0 & \ldots & 1\\
\vdots & \vdots & \vdots & \ddots & \vdots\\
0 & 0 & 0 & \ldots & 0
\end{bmatrix}
\quad
$$
Class $1$ precedes classes $2, 3, \ldots, |\calC|$; class $2$ precedes classes $3, 4, \ldots, |\calC|$; etc.
To distinguish between the previously described variant where $\mathbf{W}$ is learned and this one, we will refer to the former as the \textbf{English Learned-W model (EL)} and the latter as the \textbf{English Fixed-W model (EF)}.

\subsection{Handling Postnominal Adjectives}
\noindent In English, noun phrases consisting of a noun and one or more adjectives always place the adjectives \emph{before} the noun. 
However, this is not the case in other languages, where the adjectives can be placed before, after, or both before and after the noun. As such, we need to modify our model to accommodate such structures.

With the EL and EF models, we use a single interaction matrix $\mathbf{W}$ to score a permutation $\vpi$ of the adjectives 
$\va = \{a_1,\ldots,a_k\}$
that modifies $n$. But if we must now support adjectives both before and after the noun, we must decompose $\va$ into two sets: 
$\va^{(\mathrm{left})} = \{a^{(\mathrm{left})}_1,\ldots,a^{({\mathrm{left}})}_j\}$ and  $\va^{(\mathrm{right})} =\{a^{({\mathrm{right}})}_1,\ldots,a^{({\mathrm{right}})}_\ell\}, j \geq 2$ or $\ell \geq 2$.
Then, we can use two separate $\mathbf{W}$ matrices, $\mathbf{W}^{({\mathrm{left}})}$ and $\mathbf{W}^{({\mathrm{right}})}$, to score the adjectives that appear directly to the left and right of $n$, respectively:
\begin{multline}
    \score(\vpi, \vc) = \score(\vpi^{({\mathrm{left}})}, \vc^{({\mathrm{left}})}) \\+ \score(\vpi^{({\mathrm{right}})}, \vc^{({\mathrm{right}})})
\end{multline} 
Conveniently, maximizing $\score(\vpi, \vc)$ is equivalent to maximizing $\score(\vpi^{({\mathrm{left}})}, \vc^{({\mathrm{left}})})$ and $\score(\vpi^{({\mathrm{right}})}, \vc^{({\mathrm{right}})})$ independently.

As with English, we present two variants of the multi-lingual model, one where $\mathbf{W}^{({\mathrm{left}})}$ and $\mathbf{W}^{({\mathrm{right}})}$ are learned and one where they are fixed. We will refer to the former as the \textbf{Multi-lingual Learned-W model (ML)} and the latter as the \textbf{Multi-lingual Fixed-W model (MF)}.
The primary challenge in implementing MF is deciding what $\mathbf{W}^{({\mathrm{right}})}$ should be. While $\mathbf{W}^{({\mathrm{left}})}$ can simply be an off-upper-triangular matrix, as $\mathbf{W}$ is in EF, we need an appropriate matching $\mathbf{W}^{({\mathrm{right}})}$ that captures the different treatment given to prenominal and postnominal adjectives. Ultimately, we adopt \newcite{dixon}'s observation that postnominal adjectives follow the opposite order as do prenominal ones, and fix $\mathbf{W}^{({\mathrm{right}})}$ as a matrix with ones below the diagonal and zeroes elsewhere, i.e.~an off-lower-triangular matrix.

\subsection{Multi-lingual Word Embeddings}
In order to predict adjective order across languages, we need a joint model for word representations.
We use multi-lingual fastText \cite{fasttext} Wikipedia supervized word embeddings of dimensionality $d=300$ aligned in a single vector space (MUSE), provided by \newcite{muse}. 

\section{Data}\label{data}
This section describes our English, multi-lingual, and additional languages datasets.

\subsection{English Dataset}\label{engdata}
Multi-adjective noun phrases are surprisingly rare; analyzing 54,478 English noun phrases from the Universal Dependencies (UD) project \cite{nivre-etal-2016-universal,ud}, we find that 
only 745 of them (1.37\%) contain two or more adjectives.
As such, 
we require a large corpus to train our model. The data comprising the English dataset comes from ukWaC \cite{ukwac}, an enormous ($>$2 billion words) corpus of automatically tagged and dependency-parsed online text from the .uk domain. Unfortunately, ukWaC contains a lot of low-quality data, including non-English characters, incorrect tokenization, and part-of-speech errors.

We first extract all noun phrases where a noun is modified by multiple consecutive adjectives, i.e.~all phrases consisting of an ordered set of consecutive adjectives $[a_1,\ldots,a_k], k\ge2,$ directly preceding a noun $n$. We then disqualify all noun phrases where more than six adjectives modify a noun, because we find that such samples tend to consist of bad data,
such as ``.~.~.~.~.~.~.'' annotated as a sequence of adjectives.
Finally, MUSE fastText embeddings are only released as word--embedding dictionaries, unlike standard fastText embeddings which are built from substrings of characters. Thus, unlike conventional fastText embeddings, they are unable to infer embeddings for unseen words. And so, we need to disqualify all noun phrases which include adjectives not in these dictionaries. 

We then randomly select 12,000 phrases. Of these, 1,000 are set aside as a development set. The remaining 11,000 phrases are split in two different ways: by \textbf{token} and by \textbf{type}. Splitting by token is done by randomly picking 10,000 phrases to form the training set and letting the remaining 1,000 phrases form the testing set. Splitting by type is done by randomly picking 90\% of the unique adjective types in the data, letting all phrases where \emph{all} their adjectives belong to this 90\% form the training set, and letting the remaining phrases form the testing set. This ensures that every phrase in the testing set will contain at least one adjective not present in the training set. A summary of the English dataset can be found in \cref{englishdataset}.

\begin{table}
    \begin{center}
    \begin{tabular}{lcc}
        \toprule
        \multicolumn{3}{c}{Split by Token}\\
          & \# Phrases & \# Adj Types\\
        \midrule
        Training & 10,000 & 2,695\\
        Testing & \hphantom{1}1,000 & \hphantom{2,}806\\
        Total & 11,000 & 2,786\\
         \midrule
        \multicolumn{3}{c}{Split by Type}\\
          & \# Phrases & \# Adj Types\\
        \midrule
        Training & \hphantom{1}9,165 & 2,514\\
        Testing & \hphantom{1}1,835 & \hphantom{2,}890\\
        Total & 11,000 & 2,786\\
         \bottomrule
    \end{tabular}
    \caption{English dataset summary.} \label{englishdataset}
    \end{center}
\end{table}

\subsection{Multi-Lingual Dataset}\label{multidata}
Because our multi-lingual models are trained on multiple languages at once, we do not need as many data per language and can afford to use much smaller corpora. We obtain the non-English data used to train ML and MF from UD. UD provides treebanks with annotated dependencies in many languages, which we use to determine which adjectives are modifying which nouns. 
The English portion of this dataset re-uses the ukWaC corpus.

For each language that we choose to include, we once again extract all noun phrases where a noun is modified by multiple consecutive adjectives. This time, however, we need to account for postnominal adjectives as well. We extract all phrases where an ordered set of consecutive adjectives $[a^{({\mathrm{left}})}_1,\ldots,a^{({\mathrm{left}})}_j], j\geq2,$ precedes $n$ or an ordered set of consecutive adjectives $[a^{({\mathrm{right}})}_1,\ldots,a^{({\mathrm{right}})}_\ell], \ell \geq 2,$ follows $n$. We then once again disqualify all noun phrases which include adjectives not in the MUSE fastText dictionary. From the remaining pool, we randomly select 5,000 phrases to form our training set and 1,000 phrases to form our testing set, except for Russian, where we only have 667 phrases remaining to construct the testing set. A summary of the multi-lingual dataset can be found in \cref{multi-lingualdataset}.
\begin{table}
    \begin{center}
    \begin{tabular}{lcc}\toprule
        \multicolumn{3}{c}{Czech}\\
          & \# Phrases & \# Adj Types\\
        \midrule
        Training & 5,000 & 2,065\\
        Testing & 1,000 & \hphantom{2,}820\\
        Total & 6,000 & 2,245\\
         \midrule
        \multicolumn{3}{c}{English}\\
          & \# Phrases & \# Adj Types\\ 
        \midrule
        Training & 5,000 & 1,930\\
        Testing & 1,000 & \hphantom{2,}806\\
        Total & 6,000 & 2,092\\
         \midrule
        \multicolumn{3}{c}{German}\\
          & \# Phrases & \# Adj Types\\
        \midrule
        Training & 5,000 & 1,835\\
        Testing & 1,000 & \hphantom{2,}743\\
        Total & 6,000 & 2,040\\
         \midrule
        \multicolumn{3}{c}{Russian}\\
          & \# Phrases & \# Adj Types\\
        \midrule
        Training & 5,000 & 1,814\\
        Testing & \hphantom{2,}667 & \hphantom{2,}602\\
        Total & 5,680 & 1,920\\
         \bottomrule
    \end{tabular}
    
    \caption{Multi-lingual dataset summary.} \label{multi-lingualdataset}
    \end{center}
\end{table}

\paragraph{Criteria for Choosing Languages.}
We have two criteria for choosing languages for this dataset. Firstly, the language must have MUSE fastText embeddings, as we require embeddings aligned in a common vector space. Secondly, the UD corpora for the language must contain over 5,000 usable multi-adjective noun phrases 
to provide a sufficiently large training set.

\begin{table}
    \begin{center}
    \begin{tabular}{lcc}
        \toprule
        Language  & \# Phrases & \# Adj Types\\
        \midrule
        Bulgarian & 584 & 508\\
        Catalan & 503 & 515\\
        Croatian & 922 & 666\\
        Danish & 118 & 133\\
        Dutch & 321 & 328\\
        Estonian & 509 & 503\\
        Finnish & 250 & 254\\
        French & 621 & 612\\
        Greek & 104 & 132\\
        Hebrew & 147 & 170\\
        Hungarian & 228 & 321\\
        Italian & 397 & 419\\
        Norwegian & 756 & 543\\
        Polish & 408 & 508\\
        Portuguese & 222 & 275\\
        Slovak & 277 & 348\\
        Slovenian & 460 & 478\\
        Spanish & \llap{1,}000 & 947\\
        Swedish & 164 & 188\\
        Ukrainian & 373 & 472\\
         \bottomrule
    \end{tabular}
    \caption{Additional languages dataset summary.} \label{moredata}
    \end{center}
\end{table}
\subsection{Additional Languages Dataset}\label{adddata}
A glaring limitation of our multi-lingual dataset is that it is not typologically diverse: it contains two Germanic and two Slavic languages. Most critically, we note that in all four of its languages, adjectives predominantly precede the noun. While we are unable to train on more languages due to a lack of data, there is no reason why we cannot test on them. The additional languages dataset consists of phrases from 20 additional MUSE-supported languages using their UD corpora and the same pre-processing pipeline as described in \cref{multidata}. Among these are three Uralic languages (Estonian, Finnish, Hungarian) and one Afro-Asiatic language (Hebrew), while the rest are Indo-European. We do not include Arabic because its MUSE fastText embeddings seem to be incorrectly formatted. We also choose not to include Indonesian, Macedonian, Romanian, Turkish, or Vietnamese because they have too few ($<$50) phrases to construct a representative testing set. Meta-data describing the additional languages dataset can be found in \cref{moredata}.

\section{Experimental Details and Hyperparameters}
We split our experiments into English experiments (\cref{engexp}) and transfer learning experiments (\cref{multiexp}).
All of our models are trained for a single epoch of the relevant training data with a learning rate of 0.1 and a batch size of 32; we found a single epoch more than sufficient
for our purposes in preliminary experimentation.
We also set $|\calC|=15$ and $d=300$ for all models.
We report the exact expectation of the random baseline.
All significance testing is done with permutation tests
following \citet{dror-etal-2018-hitchhikers},
using 10,000 random permutations and significance at $\alpha=0.05$. 
All differences between model performance and the corresponding random baselines are significant with $p < 0.01$. 

\section{English Experiments}\label{engexp}
Our English experiments serve to demonstrate the basic correctness of the model. We also provide a qualitative analysis of EF.

\subsection{Predictive Accuracy}\label{engaccuracy}

We train each of the English models
on the token and type split English data
described in \cref{engdata}.
The token split allows us to evaluate the basic predictive accuracy of EL and EF,
while the type split allows us to evaluate how well the EL and EF models generalize to unseen adjective types.
Results are detailed in \cref{engresults}.
We achieve high accuracy on both the token split and type split data, demonstrating the correctness of the model. 
Importantly, our strong performance on the type split data demonstrates that EL and EF generalize well to unseen adjective types. 
We also observe that EL and EF results are similar, suggesting that adjective ordering preferences naturally tend towards a total ordering,
since learning $\mathbf{W}$ did not significantly improve results.

\begin{table}
    \begin{center}
    \begin{tabular}{l c c c} \toprule
         & EL & EF & Random\\
        \midrule
        Token split & 0.843 & 0.823 & 0.483 \\ 
        Type split & 0.836 & 0.829 & 0.482\\
         \bottomrule
    \end{tabular}
    \caption{English accuracy on different data splits. Comparing the two models on the same data split, the results do not differ significantly.} \label{engresults}
    \end{center}
\end{table}

\subsection{Validating Use of fastText}
We now address a potential confounding influence of the pre-trained fastText embeddings. We are concerned that adjective ordering information may be pre-baked into the MUSE fastText embeddings that we use, since the embeddings were trained on text where adjectives were correctly ordered. To check this, we retrain two small fastText models on a subset of 12,500 sentences from ukWaC. The first model is trained on these sentences as they are, and the second model is trained on a version of these sentences where strings of consecutive adjectives have been randomly scrambled. We then retrain the EL and EF models on the token split data with both the scrambled and unscrambled fastText vectors. Results are detailed in \cref{scrambled}.

\begin{table}
    \begin{center}
    \begin{tabular}{l c c c} \toprule
         & EL & EF & Random\\
        \midrule
        Scrambled & 0.791 & 0.797 & 0.483\\
        Unscrambled & 0.784 & 0.797 & 0.483\\
         \bottomrule
    \end{tabular}
    \caption{English accuracy with scrambled and unscrambled fastText vectors. Comparing different vectors for the same model, the results do not differ significantly.} \label{scrambled}
    \end{center}
\end{table}

That neither pair of scrambled and unscrambled results differs significantly indicates that adjective ordering information is \emph{not} coming from the fastText embeddings.
Otherwise, the unscrambled model should have outperformed the scrambled model. Due to the computational expense of retraining multi-lingual fastText, we do not repeat this validation with the multi-lingual models.

\subsection{Qualitative Evaluation of EF}
Perhaps the most convenient property of the EF model is that it is fully interpretable. We are able to, for any given adjective, extract information about which class it belongs to, and know from the model's design that classes follow a total ordering such that class 1 precedes class 2 precedes class 3, and so on. In this experiment, we first qualitatively analyze the 177 testing phrases in the token split data that EF orders incorrectly, making generalizations about what kinds of mistakes the model makes. We then make a qualitative comparison between the hierarchy that EF learns and the hierarchy proposed by \newcite{cinquebook}.

\paragraph{Types of Mistakes.} Two types of cases account for most of EF's mis-orderings. Firstly, many of the mis-ordered testing phrases deviate from typical adjective ordering tendencies because they contain adjective--noun collocates. Such phrases include ``Italian [secret service]'', ``modern [good practice]'', and ``Japanese [popular culture]'' (to illustrate how these are atypical, consider ``secret Italian meatballs'', ``good modern ethics'', and ``popular Japanese restaurant''). We note that this tends to occur with adjectives that describe \textsc{provenance}: these, while typically placed near the noun, are also often prepended to collocates. We are largely unsurprised by this, as it mirrors the intuitive observations made regarding adjective-noun collocates illustrated in \cref{ex:bear} and \cref{ex:brownbear}. An interesting direction for future work might be to model the likelihood of an adjective and a noun forming a collocate and integrate that into our current model.

Secondly, we observe that EF often mis-orders phrases containing adjectives describing $\textsc{order}$ (e.g.~``next'', ``first'', ``other'') and $\textsc{quantity}$ (e.g.~``few'', ``many''). 
Examining EF's adjective-class layer, we discover that it has placed these words together in the same class, when intuitively $\textsc{order}$ adjectives should precede $\textsc{quantity}$ adjectives (e.g.~``next few lessons'', ``first many partners''). Further experimentation would be necessary to determine why EF has done this, but we suspect intuitively that it may be because $\textsc{order}$ and $\textsc{quantity}$ adjectives are relatively small classes and are semantically similar.  If they occur more often next to other classes than next to each other, there is only a weak pressure for the model to assign these words to distinct classes.
A more rigorous error analysis would require a comprehensive dictionary of adjectives tagged with their semantic classes.\footnote{Specifically, these would have to be semantic classes comparable with those learned by EF.}
Unfortunately, constructing such a dictionary is beyond the scope of this paper.

\paragraph{Comparison with \citeauthor{cinquebook}'s Hierarchy.}
We take the 100 most common adjectives in the English dataset and use EF's adjective-class layer to determine their class memberships. We then compare these classes and their relative orderings to those proposed by \newcite{cinquebook}: $\textsc{value}$ $\rightarrow$ $\textsc{size}$ $\rightarrow$ $\textsc{shape}$ $\rightarrow$ $\textsc{color}$ $\rightarrow$ $\textsc{provenance}$.

We observe that EF follows most of \citeauthor{cinquebook}'s rules. Most notably, EF clearly learns categories of adjectives describing \textsc{size}, \textsc{color} and \textsc{provenance}, and additionally learns that \textsc{size} precedes \textsc{color} precedes \textsc{provenance}. We perform a small-scale statistical verification of this observation by hand-constructing a testing set of Cinquean phrases and using it to evaluate the similarity of EF's and \citeauthor{cinquebook}'s predictions. To do this, we first select five common adjectives from each of the five Cinquean categories. We then construct a testing set using pairs of only these $25$ adjectives based on \citeauthor{cinquebook}'s hierarchy. This gives us ${5 \choose 2}*5^2 = 250$ testing phrases. Since these are all pairs, the expected random baseline is simply 50\%. 

We then evaluate the predictive accuracy of EF on the Cinquean testing phrases. EF achieves an accuracy of 0.960 with $p<0.01$, suggesting that EF agrees with most of \citeauthor{cinquebook}'s rules. Importantly, this does \emph{not} mean that EF is 96\% accurate at ordering adjectives, but only that EF agrees with 96\% of \citeauthor{cinquebook}'s predictions on our test set. As discussed, many of EF's mistakes on real corpus data are attributable to adjective ordering exceptions like adjective-noun collocates, which \citeauthor{cinquebook}'s hierarchy does not address either.

While EF follows most of \citeauthor{cinquebook}'s \emph{existing} rules, we also observe that EF learns \emph{additional} rules not described by \citeauthor{cinquebook}. For instance, EF seems to learn a category of adjectives describing \textsc{type}, which follows adjectives describing \textsc{provenance} and contains adjectives such as ``financial'', ``technical'', and ``scientific''. This seems intuitively correct---to illustrate, consider ``Russian financial burden'', ``German technical wonder'', and ``African scientific achievement''. This suggests that an accurate adjective ordering hierarchy may need to be more complex than described by \citeauthor{cinquebook}. In particular, it seems that \citeauthor{cinquebook}'s adjective classes are too broad. An alternate interpretation is that \textsc{type} adjectives are defined by being capable of forming adjective-noun collocates with most of the nouns that they commonly modify. 

But we must emphasize that this analysis is still anecdotal.
The noted similarities and differences are difficult to quantify, and as far as we are aware there is no large-scale corpus of adjectives tagged with their Cinquean categories to enable a more reliable quantitative approach; we would ideally want such a corpus in a large number of languages. 
For now, we simply suggest that while \citeauthor{cinquebook}'s hierarchy captures many truths about adjective ordering, it does not quite grasp the entire picture.

\begin{table*}
    \begin{center}
    \begin{tabular}{lccccccc} \toprule
        &\multicolumn{3}{c}{ML (Learned $\mathbf{W}$)}
            &\multicolumn{3}{c}{MF (Fixed $\mathbf{W}$)}
            &\multirow{2}{*}{Random}\\
        \cmidrule(lr){2-4}\cmidrule(lr){5-7}
        & Transfer & Mono-ling & Joint
            & Transfer & Mono-ling & Joint
            & \\ \midrule
        Czech & $0.851$\rlap{$^{\dagger\ddagger*}$} & $0.886$\rlap{$^{\dagger*}$} & $0.899$ & $0.817$\rlap{$^{\ddagger*}$} & $0.831$\rlap{$^{\dagger*}$} & $0.888$ & $0.483$ \\
        English & $0.803$\rlap{$^{\dagger\ddagger}$} & $0.820$ & $0.820$ & $0.800$ & $0.811$ & $0.808$ & $0.487$\\
        German & $0.695$\rlap{$^{\dagger\ddagger*}$} & $0.802$ & $0.807$ & $0.732$\rlap{$^{\dagger\ddagger*}$} & $0.796$ & $0.807$ & $0.488$\\
        Russian & $0.840$\rlap{$^{\dagger\ddagger}$} & $0.893$\rlap{$^{\dagger}$} & $0.911$ & $0.859$\rlap{$^{\ddagger}$} & $0.873$\rlap{$^{\dagger}$} & $0.892$ & $0.485$\\
        \bottomrule
    \end{tabular}
    \caption{Multi-lingual accuracy. A $^{\dagger}$ denotes that a result differs significantly from the result to its right. A $^{\ddagger}$ denotes that a result differs significantly from the result two to its right. A $^*$ denotes that a result differs significantly from its ML/MF counterpart. The terminology used to describe the columns is defined in \cref{multiaccuracy}.}\label{multiresults}
    \end{center}
\end{table*}

\paragraph{Comparison with Functional Theories}
The bulk of the existing work on statistically modelling adjective ordering can be broadly separated into two categories: that which is \emph{theoretically-motivated} (e.g.~\citealp{wuff,futrell-etal-2020-determines}), and that which is \emph{empirically-motivated} (e.g.~\citealp{malouf}). The theoretically-motivated approach attempts to deduce the source of adjective ordering preferences by fitting adjective ordering data to pre-determined features derived from more fundamental functional pressures. The empirically-motivated approach attempts to fit adjective ordering data as accurately as possible by learning features from data. This paper falls into the empirically-motivated category because a hierarchical model like ours or \citeauthor{cinquebook}'s is in no way functional -- it postulates that a particular hierarchy exists but does not explain why it exists in that particular order. Importantly, this means that a hierarchical theory is not necessarily at odds with the functional theories. Rather, it is very possible that one or more functional theories might serve to explain the empirically observed hierarchies.

Interestingly, there seems to be a gap in predictive accuracy between theoretically-motivated and empirically-motivated models. For example, \newcite{wuff} and \newcite{futrell-etal-2020-determines} achieve accuracies in the low 70s, while \newcite{malouf} and this paper achieve accuracies in the 80s. While these results are hard to compare directly as they were achieved on different datasets, this suggests that there are some ordering preferences not yet captured by any existing functional theory.

\section{Transfer Learning Experiments}\label{multiexp}
An important claim of the hierarchical theory for adjective ordering is that the hierarchy applies universally across languages. If this is the case, then we should be able to accurately order adjectives from languages that we have not trained on. 

\subsection{Predictive Accuracy}\label{multiaccuracy}

We evaluate 
each of the multi-lingual models on the multi-lingual dataset in three different scenarios.
The first scenario (henceforth the \textbf{mono-lingual} scenario) addresses single-language training and testing. For this, we train one model on each of the four languages in the dataset by itself. Each model is then tested on the language that it was trained on. The second scenario (henceforth the \textbf{transfer} scenario) addresses the model's ability to generalize to unseen languages by holding out the language in question. For this, we train four models, each on every language but the one we want to test (e.g.~on Czech, English, German, but not Russian). Each model is then tested on the language that was held out during training. The third scenario  (henceforth the \textbf{joint} scenario) addresses the potential for augmenting single-language training with additional data from other languages.  For this, we train a single model on all four languages together. The model is then tested on each of the four languages individually.
Results are detailed in \cref{multiresults}.

We observe that the model performs much better than chance on the transfer scenario. This confirms the theory that universal hierarchical adjective ordering tendencies generalize across languages. Otherwise, we would expect chance level performance.
We also observe that for all languages, performance on the joint scenario is better than or equal to performance on the mono-lingual scenario, which is in turn better than or equal to performance on the transfer scenario. 
This upward trend of transfer $\leq$ mono-lingual $\leq$ joint suggests that while training on additional languages can help performance, the most important single factor is to train on the language that is being tested. In fact, given that the multi-lingual models did not achieve the same performance on the joint scenario as the English models did on the English dataset (\cref{engaccuracy}), we predict that performance on the mono-lingual scenario would have been the best for all languages if there had been more training data.
Finally, we observe that for the most part, corresponding ML and MF results do not differ significantly, suggesting once again that adjective ordering preferences tend towards a total ordering.
Taken together, these observations suggest that a universal hierarchy of adjective ordering tendencies exists, though individual languages may also feature additional unique tendencies not shared by the others.

\subsection{Testing on Additional Languages}
\begin{figure}
\begin{center}
\begin{tikzpicture}
\begin{axis}[
    xbar,
    xtick pos = right,
    ytick pos = bottom,
    bar width = 0.12cm,
    xmin = 0,
    height=11cm,
    enlarge y limits={abs=20pt},
    width=7cm,
    legend style={at={(0.5,0.83)},
      anchor=north,legend columns=-1},
    symbolic y coords={Bulgarian, Catalan, Croatian, Danish, Dutch, Estonian, Finnish, French, Greek, Hebrew, Hungarian, Italian, Norwegian, Polish, Portuguese, Slovak, Slovenian, Spanish, Swedish, Ukrainian},
    ytick={data},
    yscale = -1,
    xtick={0.2, 0.4, 0.6, 0.8, 1.0},
    ]
\addplot coordinates {(0.851,Bulgarian) (0.763,Catalan) (0.850,Croatian) (0.791,Danish) (0.819,Dutch)(0.673,Estonian)(0.702,Finnish)(0.802,French)(0.832,Greek)(0.868,Hebrew)(0.839,Hungarian)(0.740,Italian)(0.797,Norwegian)(0.779,Polish)(0.722,Portuguese)(0.770,Slovak)(0.818,Slovenian)(0.771,Spanish)(0.769,Swedish)(0.833,Ukrainian)};
\addplot coordinates {(0.487,Bulgarian) (0.494,Catalan) (0.487,Croatian) (0.492,Danish) (0.488,Dutch)(0.493,Estonian)(0.493,Finnish)(0.490,French)(0.490,Greek)(0.493,Hebrew)(0.466,Hungarian)(0.493,Italian)(0.480,Norwegian)(0.500,Polish)(0.491,Portuguese)(0.475,Slovak)(0.485,Slovenian)(0.491,Spanish)(0.492,Swedish)(0.487,Ukrainian)};
\legend{MF,Random}
\end{axis}
\end{tikzpicture}
\caption{MF accuracy on additional languages.}\label{visualize}
\end{center}
\end{figure}
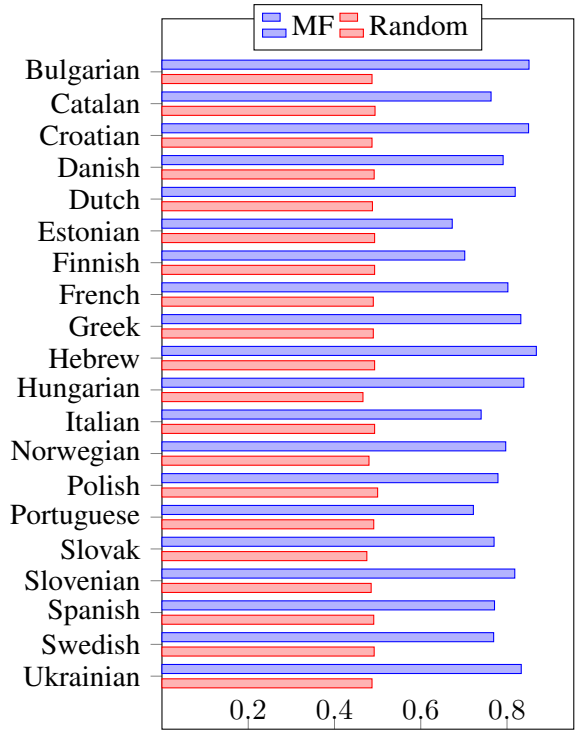

To build confidence that our findings truly generalize widely across typologically diverse languages, we train the MF model
on Czech, English, German, and Russian, and test it
on each of the languages in the additional languages dataset.
We choose to test only the MF model as the ML model would not have the data to learn a correct $\mathbf{W^{\mathrm{(right)}}}$ matrix (as Czech, English, German, and Russian tend not to have postnominal adjectives) and would thus understandably under-perform on the languages which predominantly feature postnominal adjectives (i.e.~Catalan, French, Hebrew, Italian, Portuguese, and Spanish). This experiment is conceptually identical to the multi-lingual transfer scenario. Results are detailed in \cref{moreresults} and visualized in \cref{visualize}. 

MF performs much better than chance on every language, with similar accuracies as those achieved in the transfer scenario. This gives us confidence that the conclusions drawn in \cref{multiaccuracy} do generalize widely across typologically diverse languages.

\begin{table}
    \begin{center}
    \begin{tabular}{l l c c}
        \toprule
        Language & Family & Accuracy & Random\\
        \midrule
        Bulgarian & Slavic & 0.851 & 0.487\\
        Catalan & Romance & 0.763 & 0.494\\
        Croatian & Slavic & 0.850 & 0.487\\
        Danish & Germanic & 0.791 & 0.492\\
        Dutch & Germanic & 0.819 & 0.488\\
        Estonian & Finnic & 0.673 & 0.493\\
        Finnish & Finnic & 0.702 & 0.493\\
        French & Romance & 0.802 & 0.490\\
        Greek & Greek & 0.832 & 0.490\\
        Hebrew & Semitic & 0.868 & 0.493\\
        Hungarian & Ugric & 0.839 & 0.466\\
        Italian & Romance & 0.740 & 0.493\\
        Norwegian & Germanic & 0.797 & 0.480\\
        Polish & Slavic & 0.779 & 0.500\\
        Portuguese & Romance & 0.722 & 0.491\\
        Slovak & Slavic & 0.770 & 0.475\\
        Slovenian & Slavic & 0.818 & 0.485\\
        Spanish & Romance & 0.771 & 0.491\\
        Swedish & Germanic & 0.769 & 0.492\\
        Ukrainian & Slavic & 0.833 & 0.487\\
         \bottomrule
    \end{tabular}
    \caption{MF accuracy on additional languages.} \label{moreresults}
    \end{center}
\end{table}

\section{Conclusion}
We built an interpretable, multi-lingual latent-variable model of hierarchical adjective ordering that directly enforces a hierarchy of semantic classes and is trained entirely using corpus data. We found that our fixed-W variants, which enforce total orderings of semantic classes, perform similarly to our learned-W variants, suggesting that adjective ordering preferences naturally tend towards total orderings. We also found that our model is able to accurately order adjectives from 24 different languages, regardless of whether it was directly trained on them, although it does benefit from having been trained on the language on which it is tested. Interestingly, we were able to achieve high predictive accuracy on languages predominantly featuring postnominal adjectives (e.g.~French, Spanish), despite having only trained on languages predominantly featuring prenominal ones (Czech, English, German, Russian), by simply reversing the prenominal adjective ordering rules for postnominal ones. 

In summary, our work presents converging evidence that adjectives exhibit universal hierarchical ordering tendencies, with the added observations that individual languages feature additional unique tendencies not shared by others, and that adjective ordering is symmetric with respect to the noun.

\bibliographystyle{acl_natbib}
\bibliography{anthology,emnlp2020}

\newpage
\clearpage
\appendix

\section{Reproducibility}
In the interest of fostering reproducibility, we provide the following additional information about our data, models, and computing infrastructure.
\subsection{Data}
We use Universal Dependencies (UD) 2.5 and ukWaC, which can be found at \url{http://hdl.handle.net/11234/1-3105} and \url{https://wacky.sslmit.unibo.it/doku.php}, respectively. Note that UD has since been updated to version 2.6.
\subsection{Model Parameters and Runtime}
The learned-W models (EL, ML) have $2*|\calC|^2+d*|\calC|=4,950$ learnable parameters. The fixed-W models (EF, MF) have $d*|\calC|=4,500$ learnable parameters. The time taken to train each model varies based on the number of training samples---as a rule of thumb, training the learned models takes about 1.5-2 hours per 10,000 samples, while training the fixed models takes about 1 hour per 10,000 samples. Training all of the model variants necessary to reproduce this paper in full takes about 24 hours. Testing either model type takes only several minutes per 1,000 samples.
\subsection{Computing Infrastructure}
All our development, training, and testing was done on a personal computer with the following specifications:
\begin{itemize}
    \item Operating System: Windows 10 Pro (64-bit)
    \item CPU: Intel Core i7-7700k @ 4.20 GHz
    \item GPU: None
    \item RAM: 64GB DDR4
    \item Storage Used: Approximately 200GB
\end{itemize}
\subsection{Other Notes}
We did not use validation sets as we saw little value to extensively tuning the model, since we were trying to explore the properties of a natural phenomenon rather than aiming to achieve the highest possible accuracy. All reported results are from the first time each model variant was tested.

\end{document}